\documentclass[sigconf]{acmart}

\usepackage{bm}
\usepackage{float}
\usepackage{algorithm}
\usepackage{algorithmic}
\usepackage{balance}

\AtBeginDocument{%
 }

\copyrightyear{2026}
\acmYear{2026}
\setcopyright{cc}
\setcctype{by}
\acmConference[KDD '26]{Proceedings of the 32nd ACM SIGKDD Conference on Knowledge Discovery and Data Mining V.2}{August 09--13, 2026}{Jeju Island, Republic of Korea}
\acmBooktitle{Proceedings of the 32nd ACM SIGKDD Conference on Knowledge Discovery and Data Mining V.2 (KDD '26), August 09--13, 2026, Jeju Island, Republic of Korea}
\acmDOI{10.1145/3770855.3819059}
\acmISBN{979-8-4007-2259-2/2026/08}

\newcommand{\newtext}[1]{\textcolor{blue}{#1}}
\newcommand{\modtext}[1]{\textcolor{purple}{#1}}

\newenvironment{newsection}{\color{blue}}{}
\newenvironment{modsection}{\color{purple}}{}

\renewcommand{\newtext}[1]{#1}
\renewcommand{\modtext}[1]{#1}
\renewenvironment{newsection}{}{}
\renewenvironment{modsection}{}{}

\begin{document}

\title{Stabilizing Physics-Informed Consistency Models via Structure-Preserving Training}

\author{Che-Chia Chang}
\affiliation{%
  \department{Institute of Artificial Intelligence Innovation}
  \institution{National Yang Ming Chiao Tung University}
  \city{Hsinchu}
  \country{Taiwan}
}
\email{ccchang.c@nycu.edu.tw}

\author{Chen-Yang Dai}
\affiliation{%
  \department{Department of Applied Mathematics}
  \institution{National Yang Ming Chiao Tung University}
  \city{Hsinchu}
  \country{Taiwan}
}
\email{cydai.sc12@nycu.edu.tw}

\author{Te-Sheng Lin}
\affiliation{%
  \department{Department of Applied Mathematics}
  \institution{National Yang Ming Chiao Tung University}
  \city{Hsinchu}
  \country{Taiwan}
}
\affiliation{%
  \department{National Center for Theoretical Sciences}
  \institution{National Taiwan University}
  \city{Taipei}
  \country{Taiwan}
}
\email{teshenglin@nycu.edu.tw}

\author{Ming-Chih Lai}
\affiliation{%
  \department{Department of Applied Mathematics}
  \institution{National Yang Ming Chiao Tung University}
  \city{Hsinchu}
  \country{Taiwan}
}
\email{mclai@math.nctu.edu.tw}

\author{Chieh-Hsin Lai}
\affiliation{%
  \department{Department of Applied Mathematics}
  \institution{National Yang Ming Chiao Tung University}
  \city{Hsinchu}
  \country{Taiwan}
}
\email{chiehhsinlai@gmail.com}


\begin{abstract}
  We propose a physics-informed consistency modeling framework for solving partial differential equations (PDEs) via fast, few-step generative inference. We identify a key stability challenge in physics-constrained consistency training, where PDE residuals can drive the model toward trivial or degenerate solutions, degrading the learned data distribution. To address this, we introduce a structure-preserving two-stage training strategy that decouples distribution learning from physics enforcement by freezing the coefficient decoder during physics-informed fine-tuning. We further propose a two-step residual objective that enforces physical consistency on refined, structurally valid generative trajectories rather than noisy single-step predictions. The resulting framework enables stable, high-fidelity inference for both unconditional generation and forward problems. We demonstrate that forward solutions can be obtained via a projection-based zero-shot inpainting procedure, achieving consistent accuracy of diffusion baselines with orders of magnitude reduction in computational cost. 
\end{abstract}




\maketitle

\section{Introduction}

Diffusion Models~\cite{sohl2015deep,song2019generative,ho2020denoising,song2021scorebased,lai2025principles} have achieved remarkable success in high-fidelity generation of images and audio. Motivated by their ability to model complex, high-dimensional distributions, recent works have adapted diffusion models to the domain of scientific machine learning~\cite{SHU2023111972,doi:10.1137/24M1636071}. Unlike traditional numerical solvers or Physics-Informed Neural Networks (PINNs)~\cite{RAISSI2019686}, which determine a single solution to a given PDE, these generative models learn the full distribution of possible solutions. This capability is crucial for ill-posed inverse problems, such as reconstructing complex physical fields from sparse observations~\cite{diffusionpde}, where a single deterministic prediction fails to capture the multiplicity of valid states.

However, despite their expressive power, diffusion models suffer from iterative sampling procedures that require hundreds or even thousands of neural network evaluations to generate a single sample. This limits their practicality for scientific applications that demand real-time or large-scale simulations. To overcome this bottleneck, Consistency Models (CMs) \cite{song2023consistency} have been proposed as a fast alternative, enabling one- or few-step generation from noise to data.

While Consistency Models offer a promising path toward real-time scientific simulation, adapting them to strictly enforce physical laws introduces additional optimization challenges. Unlike standard image generation, where perceptual quality is the primary metric, scientific solutions must adhere to rigorous governing equations. Empirically, we observe that directly incorporating Partial Differential Equation (PDE) residuals into the consistency objective often leads to unstable training dynamics and suboptimal convergence. One common difficulty in this setting is that physics losses may admit low-complexity or degenerate solutions, such as near-zero or overly smooth fields, that yield small residual values but fail to represent meaningful physical states. In practice, this often manifests as severe mode collapse, where consistency training converges to a narrow subset of admissible solutions despite satisfying the imposed physical constraints. This mismatch between physical constraints and the self-consistency objective motivates the need for specialized training and inference strategies to obtain both distributional fidelity and physical correctness.

In this work, we propose sCM-PINN, a physics-informed consistency modeling framework for efficient inference in forward and conditional PDE problems. Our approach introduces a two-stage training strategy that decouples distribution learning from physics-constrained fine-tuning, thereby stabilizing optimization in the presence of sensitive residual losses. To prevent the physical constraints from degrading the learned data manifold, we partition the model output into coefficient and solution channels, freezing the coefficient decoder during the physics-informed phase. Furthermore, to bridge the gap between consistency training and physical validity, we introduce a two-step residual-based constraint that enforces PDE consistency across refined trajectories rather than noisy, single-step predictions. Together, these design choices enable fast, reliable inference for both unconditional sampling and forward problem solving.

Our main contributions are:
\begin{itemize}
    \item Two-Stage Training Framework: A stabilized optimization pipeline for physics-informed consistency models that prevents gradient interference between data and physics objectives.
    \item Channel-Partitioned Architecture: A strategy to preserve the integrity of the learned data distribution by isolating and freezing the coefficient manifold.
    \item Two-Step PDE Residual Constraint: A robust physical supervision mechanism that aligns the consistency sampling trajectory with the underlying PDE operator.
    \item Projection-Based Inference: A zero-shot inpainting formulation for forward PDE problems that bypasses the need for expensive test-time gradient descent.
    \item Physical Evaluation Metric: The introduction of the relative $H^1$ error metric to rigorously quantify the alignment of generated samples with the PDE, ensuring consistency in both solution values and their spatial derivatives.
\end{itemize}

Finally, for reproducibility and future research, we release our complete codebase at: \url{https://github.com/twMisc/sCM-PINN}.

\section{Preliminaries and Related Work}
\label{sec:preliminaries}

\subsection{Generative Modeling of PDE Solutions}
\label{subsec:gen_pde}

Conventional learning-based approaches for scientific computing, such as Neural Operators~\citep{li2021fourier,osti_2281727}, typically formulate PDE solving as learning a deterministic mapping from coefficient fields to solution fields. This formulation is highly effective for well-posed forward problems, but becomes fundamentally limited in inverse or ill-posed settings, where solutions may be non-unique, and observations are sparse or incomplete.

To address these limitations, recent work has reformulated PDE solving as a \emph{generative modeling} problem. Methods such as DiffusionPDE~\citep{diffusionpde} aim to learn the joint distribution $p(\mathbf{a}, \mathbf{u})$ over physical coefficients $\mathbf{a}$ and state variables $\mathbf{u}$. Let $\mathbf{x} = [\mathbf{a}, \mathbf{u}] \in \mathbb{R}^D$ denote the concatenated system state. By learning the score function $\nabla_{\mathbf{x}} \log p(\mathbf{x})$, these models unify forward and inverse problems within a single conditional sampling framework: forward prediction corresponds to sampling from $p(\mathbf{u} \mid \mathbf{a})$, while inverse inference corresponds to sampling from $p(\mathbf{a} \mid \mathbf{u})$. However, such score-based approaches typically rely on iterative sampling procedures, leading to high inference costs in high-dimensional PDE settings.

\subsection{Consistency Models for Fast Sampling}
\label{subsec:cm_background}

Consistency Models~\citep{song2023consistency} are a class of generative models designed to overcome the inference inefficiency of diffusion-based sampling. Instead of iteratively integrating a reverse-time Stochastic differential equation or Probability Flow ODE (PF-ODE)~\citep{song2021scorebased}, CMs learn a continuous consistency function $\bm{f}_\theta(\mathbf{x}_t, t)$ that maps any point $\mathbf{x}_t$ along the PF-ODE trajectory directly to its corresponding data sample $\mathbf{x}_0$. The training objective enforces a \emph{self-consistency} condition,
\begin{equation}
\bm{f}_\theta(\mathbf{x}_t, t) = \bm{f}_\theta(\mathbf{x}_{t'}, t'),
\quad \forall t, t' \in [0, T],
\end{equation}
ensuring that predictions remain invariant across the trajectory. By distilling the iterative integration process into a single mapping, CMs enable one-step or few-step generative inference. In this work, we adopt the continuous-time formulation (sCM)~\citep{lu2025simplifying}, which eliminates discretization error present in discrete-time consistency training and improves training stability.

\subsection{Physics-Informed Generative Models}
\label{subsec:physics_guidance}

While consistency models (CMs) provide an efficient framework for few-step generative inference, they lack the inherent inductive biases required to satisfy the underlying governing equations of physical systems. Bridging this gap necessitates the integration of PDE constraints directly into the CM training and sampling trajectories. Within the broader literature of generative modeling for PDEs, this integration typically follows two distinct paradigms: inference-time guidance, which steers the sampling process toward physical validity, and training-time supervision, which embeds physical laws into the model’s learned parameters via residual-based loss functions.

\paragraph{Inference-Time Guidance}
Inference-time guidance methods, such as Diffusion Posterior Sampling (DPS)~\citep{chung2023diffusion}, enforce physical constraints during the sampling process by modifying the score function with gradients derived from the PDE residual operator $\mathcal{R}(\cdot)$. A typical formulation is
\begin{equation}
  \nabla_{\mathbf{x}_t} \log p(\mathbf{x}_t \mid \mathcal{R}=0)
  \approx
  \nabla_{\mathbf{x}_t} \log p(\mathbf{x}_t)
  - \eta \nabla_{\mathbf{x}_t}
  \left\| \mathcal{R}\big(\hat{\mathbf{x}}_0(\mathbf{x}_t)\big) \right\|^2.
\end{equation}
While flexible and model-agnostic, this approach incurs substantial computational overhead, as it requires backpropagation through both the neural network and the differential operator at each sampling step. Moreover, gradients of PDE residuals can be poorly conditioned, often necessitating careful tuning of guidance strengths across different problem settings.

\paragraph{Training-Time Supervision}

An alternative paradigm incorporates physics constraints directly into the
training objective. Physics-Informed Diffusion Models (PIDMs)~\citep{bastek2025physicsinformed} optimize a composite loss that combines the denoising objective with a PDE
residual penalty:
\begin{equation}
  \mathcal{L}_{\text{PIDM}}(\theta) =
  \mathbb{E}_{t, \mathbf{x}_0}
  \left[
    \ell_{\text{denoise}}(\theta)
    + \lambda \left\| \mathcal{R}\big(\hat{\mathbf{x}}_0^\theta\big) \right\|^2
    \right].
\end{equation}
By amortizing physics enforcement into training, this strategy enables fast, physics-consistent generation at inference time. However, extending this paradigm to consistency models presents additional challenges. As discussed in Section~\ref{sec:method}, the consistency objective can be sensitive to gradient variance introduced by PDE residual terms, motivating the stabilized training strategy proposed in this work.

\section{Methodology}
\label{sec:method}

We present a physics-informed consistency training framework designed to bridge the gap between fast generative inference and rigorous physical validity for forward PDE problems. Our approach builds upon the sCM framework~\citep{lu2025simplifying}. In incorporating physics-based constraints, we identify a common optimization failure mode in which the model collapses toward degenerate solutions that minimize the PDE residual but fail to preserve the target data distribution. To address this issue, we propose a training strategy that combines a two-stage optimization protocol with a structure-preserving channel partition.

\subsection{Continuous-Time Consistency Parameterization}
\label{subsec:trigflow_param}

We adopt the TrigFlow parameterization introduced in the sCM framework~\citep{lu2025simplifying}. This formulation learns a consistency function $\bm{f}_\theta(\mathbf{x}_t, t)$ that maps points along the probability flow trajectory directly to the corresponding clean data sample $\mathbf{x}_0$.

Let the state $\mathbf{x} = [\mathbf{a}, \mathbf{u}]$ represent the concatenation of the coefficient field $\mathbf{a}$ and the solution field $\mathbf{u}$. For a state $\mathbf{x}_t$ at time $t \in [0, \pi/2]$, the model is defined as:
\begin{equation}
  \bm{f}_\theta(\mathbf{x}_t, t) = \cos(t) \mathbf{x}_t - \sin(t) \sigma_d F_\theta\!\left(\frac{\mathbf{x}_t}{\sigma_d}, t\right),
  \label{eq:trigflow}
\end{equation}
where $F_\theta$ is a neural network parameterizing the normalized velocity field and $\sigma_d$ denotes the data standard deviation. This parameterization smoothly interpolates between the data distribution (at $t=0$) and a Gaussian prior (at $t=\pi/2$), and serves as a numerically stable backbone for incorporating physical constraints.

\subsection{Stabilizing Physics-Constrained Consistency Training}
\label{subsec:Stabilizing}

\begin{figure}[t]
  \centering
  \includegraphics[width=\linewidth]{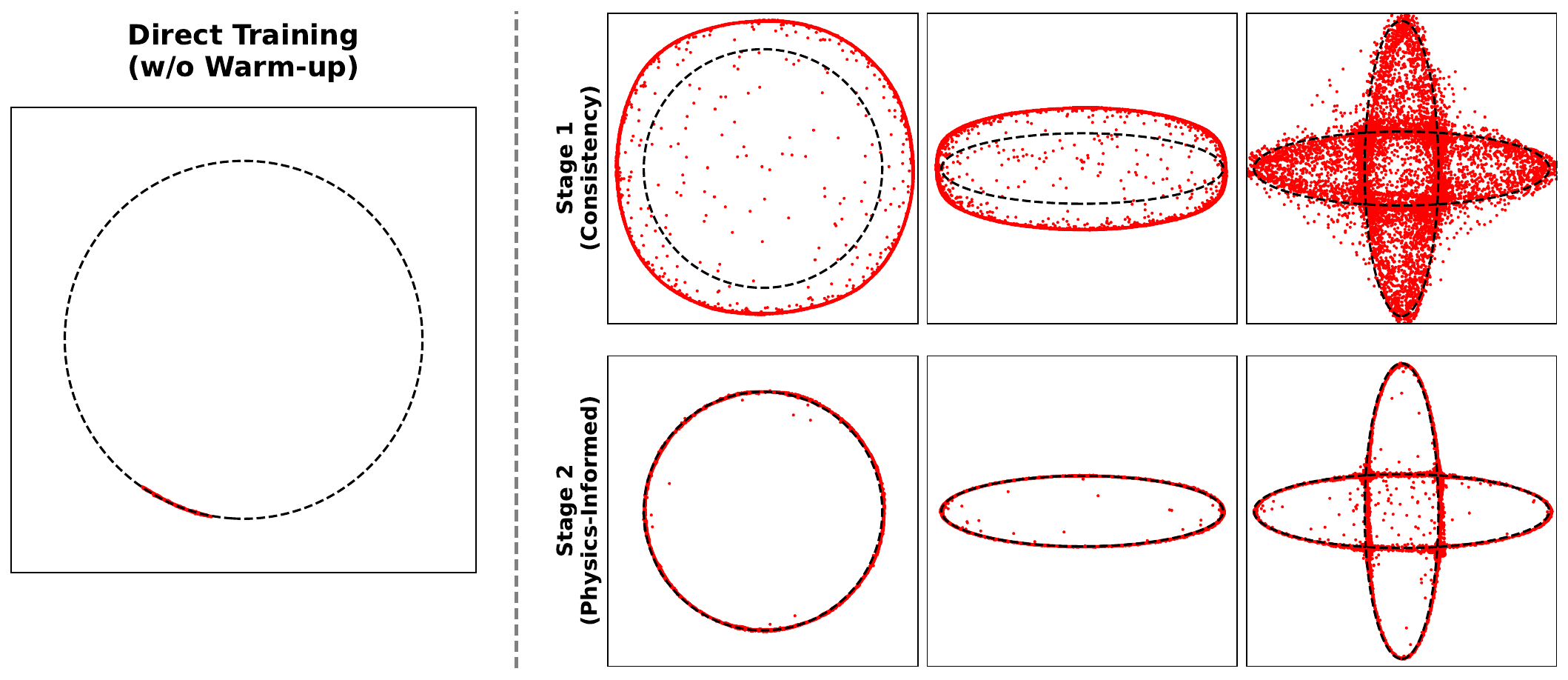}
  \caption{\textbf{Necessity of Two-Stage Training.} We visualize samples generated by physics-constrained consistency training on 2D toy manifolds (Circle, Ellipse, Double Ellipse). 
  \textbf{(Left)} Directly enforcing constraints during consistency training leads to unstable dynamics and severe mode collapse, covering only a small fraction of the target manifold.
  \textbf{(Right)} Our two-stage protocol first learns a stable, high-coverage distribution (Stage 1), then refines samples to satisfy geometric constraints (Stage 2), achieving both diversity and physical validity.
  }
  \label{fig:toy_consistency}
\end{figure}

Integrating PDE constraints into consistency models introduces significant optimization challenges. In particular, directly optimizing a composite objective from random initialization often leads to mode collapse or convergence to trivial solutions. As illustrated in Figure~\ref{fig:toy_consistency} (Left), a model trained directly on the physics objective fails to capture the global topology of the manifold, learning only a partial segment of the solution space.

To address this instability, we adopt a two-stage training protocol, validated in preliminary studies~\citep{chang2025consistency}:
\begin{enumerate}
  \item \textbf{Stage 1 (Distribution Learning):} The model is first trained to approximate the data distribution $p(\mathbf{a}, \mathbf{u})$ without explicit physical constraints. We utilize the standard adaptive sCM objective~\citep{lu2025simplifying}, which minimizes the variance of the consistency update. The loss is defined as:
        \begin{equation}
          \mathcal{L}_{\text{sCM}}(\theta, \phi) = \mathbb{E}_{\mathbf{x}_0, \mathbf{z}, t} \left[
            \frac{e^{w_\phi(t)}}{D} \left\|
            F_\theta\left(\tfrac{\mathbf{x}_t}{\sigma_d}, t\right) - \mathbf{y}_t
            \right\|_2^2 - w_\phi(t)
            \right],
          \label{eq:scm_loss_stage1}
        \end{equation}
        where $w_\phi(t)$ is a learnable weighting function and $\mathbf{y}_t$ is the target estimate computed using the stop-gradient parameters $\theta^-$. Crucially, $\mathbf{y}_t$ includes a first-order correction based on the analytic tangent $\dot{\mathbf{x}}_t = \cos(t)\mathbf{z} - \sin(t)\mathbf{x}_0$:
        \begin{equation}
          \mathbf{y}_t \triangleq F_{\theta^-}\left(\tfrac{\mathbf{x}_t}{\sigma_d}, t\right) + \cos(t) \left[ \frac{\partial \bm{f}_{\theta^-}}{\partial t} + \nabla_{\mathbf{x}} \bm{f}_{\theta^-} \cdot \dot{\mathbf{x}}_t \right].
        \end{equation}
        This stage acts as a "warm-up," establishing a robust mapping for the joint distribution of coefficients and solutions before physical constraints are applied.

  \item \textbf{Stage 2 (Physics-Informed Fine-tuning):} Once the generative backbone is established, the model is fine-tuned using the composite physics loss. In this stage, we freeze the coefficient decoder (as detailed in Sec.~\ref{subsec:channel_partition}) and simplify the consistency term to focus on projecting generated samples onto the valid physical manifold.
\end{enumerate}

\paragraph{Optimization Bias and Mode Collapse}
While Stage~1 establishes a valid data distribution, the introduction of the physics loss in Stage~2 creates a new optimization landscape that often favors degenerate solutions. The PDE residual $\mathcal{R}(\mathbf{u}, \mathbf{a})$ is distribution-agnostic; it is minimized by \textit{any} valid pair, including low-complexity states (e.g., constant coefficient fields) that are topologically simpler than the ground truth data. 
Consequently, unconstrained gradient descent tends to shift the coefficient field $\mathbf{a}$ toward these homogeneous configurations to minimize the loss rapidly, effectively disregarding the distribution learned in the warm-up phase. This collapse, which leads to the generation of trivial solutions (as demonstrated in our ablation study in Section~\ref{subsec:ablation_decoder}), necessitates the structural constraints introduced below.

\subsection{Structure-Preserving Channel Partitioning}
\label{subsec:channel_partition}

\begin{figure}[t]
    \centering
    \includegraphics[width=\linewidth]{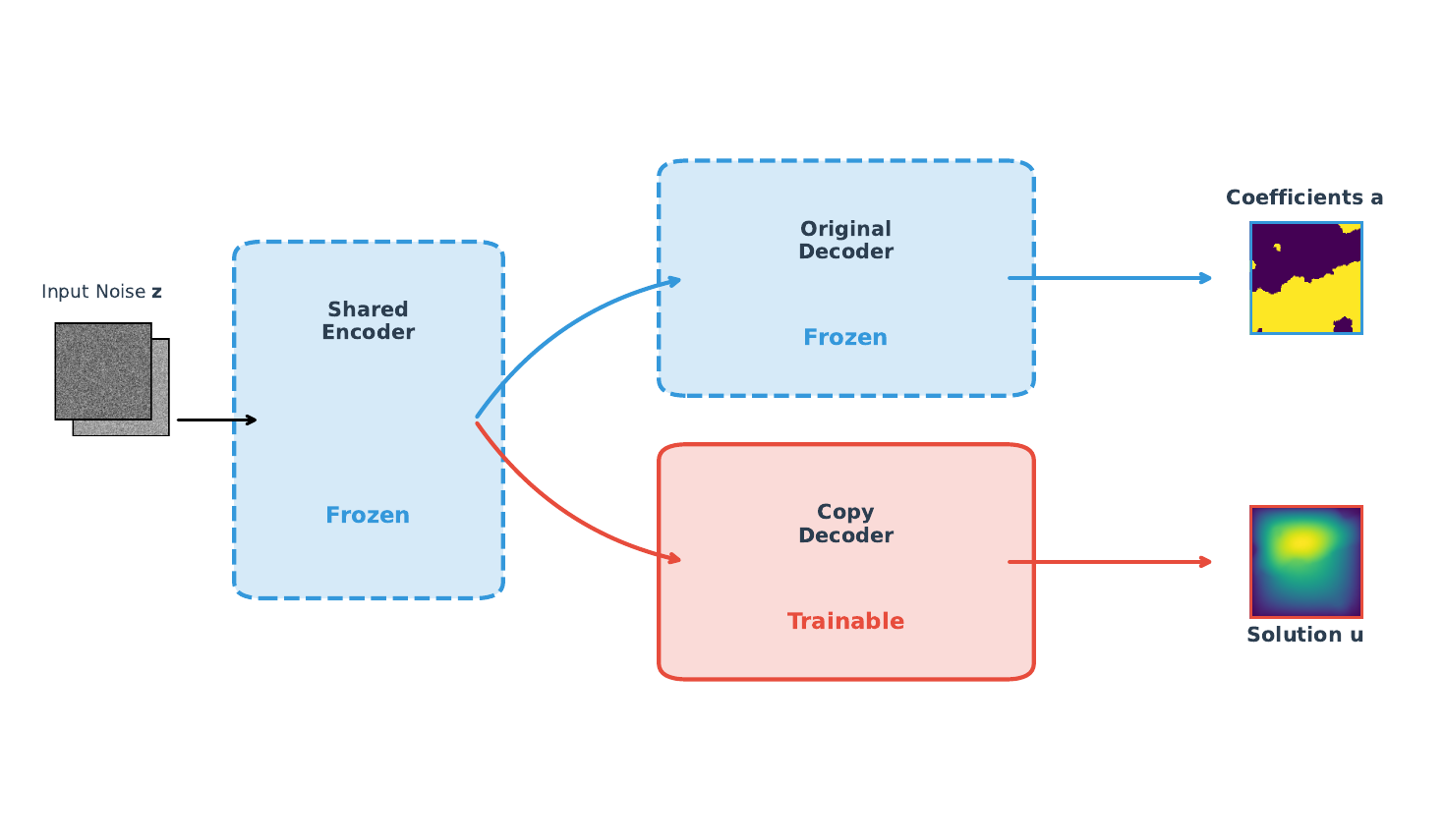}
    \caption{\textbf{Architecture of the Structure-Preserving Consistency Model.} To mitigate coefficient mode collapse during physics-informed fine-tuning, we employ a split-decoder architecture. The \textbf{Frozen Backbone} (blue, dashed) consists of the shared encoder and the Stage~1 decoder, whose parameters are fixed to preserve the learned coefficient distribution $\mathbf{a}$. The \textbf{Active Branch} (red, solid) is a trainable decoder that generates the solution field $\mathbf{u}$ conditioned on the fixed latent representation. The final output is the joint state $\mathbf{x} = (\mathbf{a}, \mathbf{u})$.}
    \label{fig:architecture}
\end{figure}

To enforce the coefficient distribution learned in Stage 1, we introduce a split-decoder architecture that aligns the network's capacity with the causal structure of the forward problem. As illustrated in Figure~\ref{fig:architecture}, our model modifies the standard U-Net backbone by duplicating the decoder branch while sharing a single encoder.

\paragraph{Architecture Design} 
The network processes the input $\mathbf{x}_t$ through a shared encoder and bifurcates into two parallel decoding paths:

\begin{itemize}
    \item \textbf{The Frozen Backbone (Blue):} This path comprises the shared encoder and the \textit{Original Decoder}. These modules are initialized with pre-trained weights from Stage 1 and are strictly frozen. This branch is responsible solely for reconstructing the coefficient field $\mathbf{a}$. By locking these weights, we ensure that the consistency mapping for the problem definition remains immutable, effectively preventing the optimizer from deviating from the valid coefficient distribution.
    
    \item \textbf{The Active Branch (Red):} We introduce a \textit{Copy Decoder} that duplicates the architecture of the Stage 1 decoder. This branch is fully trainable and focuses exclusively on generating the solution field $\mathbf{u}$. It receives the fixed latent embeddings from the frozen encoder and minimizes the physics residuals to generate a solution $\mathbf{u}$ that is physically consistent with the coefficients.
\end{itemize}

\paragraph{Forward Pass} 
During the forward pass, the network output is partitioned into coefficient channels $F_\theta^{(a)}$ and solution channels $F_\theta^{(u)}$. We discard the solution channels from the frozen branch and the coefficient channels from the active branch. The final generated state is constructed by concatenating the valid problem parameters from the frozen path with the optimized solution from the active path:
\begin{equation}
    F_\theta(\mathbf{x}_t, t) = \text{Concat}\left( 
    \underbrace{F_{\theta_{\text{frozen}}}^{(a)}(\mathbf{x}_t, t)}_{\text{Frozen}}, \,\,
    \underbrace{F_{\theta_{\text{active}}}^{(u)}(\mathbf{x}_t, t)}_{\text{Trainable}} 
    \right).
\end{equation}
This structural constraint guarantees that the generated pair $(\mathbf{a}, \mathbf{u})$ is physically aligned, as the solution $\mathbf{u}$ is optimized specifically for the fixed coefficient topology defined by $\mathbf{a}$.

\subsection{Two-Step Consistency Sampling Operator}
\label{subsec:two_step}

A naive approach to physics-informed consistency training applies the PDE residual directly to single-step predictions generated from pure noise. Empirically, however, we observe that these raw one-step samples lack spatial smoothness and exhibit significantly higher PDE residuals compared to refined solutions. Because differential operators are highly sensitive to high-frequency artifacts, minimizing physics losses on these rough approximations produces noisy gradients that destabilize physics-informed consistency training.

To mitigate this, we introduce the \textit{Two-Step Consistency Sampling Operator}, which evaluates physics constraints after a single refinement step rather than on raw one-step samples. By performing a single re-noising and denoising step, this operator projects the sample closer to the smooth solution manifold before evaluating the physics loss. This ensures that the PDE constraints are enforced on structurally valid candidates rather than noisy artifacts.

\begin{definition}[Two-Step Consistency Sampling Operator]
  Let $\bm{f}_\theta(\mathbf{x}, t)$ be the consistency model. For initial noise $\mathbf{z}$ and auxiliary noise $\mathbf{z}'$, we define the two-step operator $\hat{\bm{f}}_\theta$ as:
  \begin{equation}
    \hat{\bm{f}}_\theta(\mathbf{z}, \mathbf{z}'; T, t') \triangleq \bm{f}_\theta\left( \underbrace{ \sin(t') \mathbf{z}' + \cos(t') \bm{f}_\theta(\mathbf{z}, T) }_{\text{renoised state } \mathbf{x}_{t'}}, \; t' \right).
  \end{equation}
\end{definition}

This operator first produces a coarse estimate using $\bm{f}_\theta(\mathbf{z}, T)$, then re-injects noise and applies the consistency model again at an intermediate time $t'$. Compared to one-step sampling, this procedure yields intermediate states that are empirically closer to the data manifold while remaining stochastic. As a result, PDE residuals evaluated on $\hat{\bm{f}}_\theta$ exhibit reduced gradient variance, which in practice leads to more stable and effective physics-informed training.

We next incorporate this operator into a physics-informed Stage-2 objective that balances consistency preservation with robust enforcement of PDE constraints.

\begin{newsection}
In contrast to approaches such as PBFM~\cite{baldan2026physics} that require memory-intensive multi-step ODE unrolling to mitigate discretization errors, our model natively maps to the solution manifold, bounding evaluations without unrolling bottlenecks.
\end{newsection}

\subsection{Physics-Informed Stage 2 Objective}
\label{subsec:objective}

Our Stage~2 training objective fine-tunes the model to satisfy physical constraints while preserving the generative consistency learned in Stage~1. Unlike the adaptive variance minimization used in Stage~1, we employ an unweighted consistency loss to serve as a stable reference during physics optimization, avoiding the risk that adaptive weights suppress consistency precisely when physics gradients are strongest.

\paragraph{Unweighted Consistency Term ($\ell_{\text{sCM-L2}}$)}
We define the Stage~2 consistency loss as the squared Euclidean error between the model prediction and the consistency target. Consistent with our split-architecture, this loss is applied exclusively to the solution channels $\mathbf{u}$:
\begin{equation}
  \ell_{\text{sCM-L2}}(\theta; \mathbf{x}_t, t) = \left\|
  F_\theta^{(u)}\!\left(\tfrac{\mathbf{x}_t}{\sigma_d}, t\right) - \mathbf{y}_t^{(u)} \right\|_2^2,
  \label{eq:scm_loss_stage2}
\end{equation}
where $\mathbf{y}_t^{(u)}$ corresponds to the solution channels of the target estimate defined in Eq.~\eqref{eq:scm_loss_stage1}. By removing the adaptive weighting terms, we enforce a uniform consistency constraint, preventing the weighting function from diminishing the loss contribution at time steps where physical constraints are most active.

\paragraph{Total Objective and Self-Adaptive Balancing}
We integrate this consistency regularization with the physics constraints. To manage the gradient conflict between the consistency loss and the PDE residuals, we employ the Self-Adaptive weighting strategy (SA-PINN)~\citep{MCCLENNY2023111722} to automatically balance competing objectives.

We introduce learnable mask parameters $\bm{\lambda} = \{\lambda_0, \lambda_1, \lambda_2\}$ and formulate the training as a Min-Max game. Importantly, for computational efficiency and tighter gradient coupling, the consistency term and the single-step PDE term share the same sampling trajectory:
\begin{modsection}
\begin{align}
  \mathcal{L}_{\text{total}}(\theta, \bm{\lambda}) =
   & \quad \mathbb{E}_{\mathbf{x}_0, \mathbf{z}, t} \Bigg[
    \underbrace{ \sigma(\lambda_0) \ell_{\text{sCM-L2}}(\theta; \mathbf{x}_t, t) }_{\text{Consistency Loss}}
    + \underbrace{ \sigma(\lambda_1) \left\|
    \mathcal{R}\big(\bm{f}_\theta(\mathbf{x}_t, t)\big) \right\|_p^p }_{\text{Single-Step PDE Residual}}
  \Bigg] \nonumber                                         \\
   & + \mathbb{E}_{\mathbf{z}, \mathbf{z}', t'} \Bigg[
    \underbrace{ \sigma(\lambda_2) \left\|
    \mathcal{R}\big(\hat{\bm{f}}_\theta(\mathbf{z}, \mathbf{z}'; T, t')\big) \right\|_p^p }_{\text{Two-Step Robustness}}
    \Bigg],
  \label{eq:total_loss}
\end{align}
where $\sigma(\lambda) = \frac{1}{1+e^{-\lambda}}$ is the sigmoid gating function, and $\| \cdot \|_p$ denotes the $L^p$ norm. 
\end{modsection}
The single-step term preserves local physical fidelity along the consistency trajectory, while the two-step term enforces robustness on denoised, structurally valid samples.

\paragraph{Optimization Dynamics}
The training follows an adversarial update rule:
\begin{equation}
  \theta \leftarrow \theta - \eta_\theta \nabla_\theta \mathcal{L}_{\text{total}}, \quad
  \lambda_i \leftarrow \lambda_i + \eta_\lambda \nabla_{\lambda_i} \mathcal{L}_{\text{total}}.
\end{equation}
This mechanism allows the loss weights to dynamically adapt to the training dynamics. For instance, if the model begins to drift from the consistency manifold (increasing $\ell_{\text{sCM-L2}}$) to satisfy the physics, the gradient ascent step increases $\sigma(\lambda_0)$, forcing the model to re-prioritize consistency in subsequent updates.

\begin{newsection}
Because the scalars $\lambda_i$ update at the end of the computational graph, SA-PINN requires zero extra backward passes. This structural resolution bypasses the massive computational overhead of gradient-conflict methods like ConFIG~\cite{liu2025config}, which can triple memory usage and training time.
\end{newsection}

\begin{newsection}

\paragraph{Theoretical Motivation for the Two-Step Residual.}
    
Our motivation for the two-step residual is structural, not only empirical. Let $\mathbf x$ be the first-step prediction, and define the refined second-step sample
\begin{equation}
\mathbf{Y}_{\mathbf{x},t'} := \boldsymbol{f}_\theta(a_{t'}\mathbf{x} + b_{t'}\mathbf{z}, t'), \qquad \mathbf{z} \sim \mathcal{N}(\mathbf{0},\mathbf{I}).
\end{equation}
where $a_{t'}$ and $b_{t'}$ are the scalar coefficients defined by the TrigFlow schedule.
If we consider the $L^2$ penalty ($p=2$), the conditional two-step physics loss is
\begin{equation}
\mathcal{L}_{2}(\mathbf{x},t') := \mathbb{E}_{\mathbf{z}}\big[\|\mathcal{R}(\mathbf{Y}_{\mathbf{x},t'})\|^{2}\big].
\end{equation}
It admits the exact decomposition
\begin{equation}
\mathcal L_{2}(\mathbf x,t') =
\Big\|\mathbb E_{\mathbf z}[\mathcal R(\mathbf Y_{\mathbf x,t'})]\Big\|^{2} + 
\operatorname{tr}(\operatorname{Cov_{\mathbf z}} (\mathcal R(\mathbf Y_{\mathbf x,t'})) ).
\end{equation}
Thus, unlike the one-step loss $\|\mathcal R(\mathbf x)\|^2$, the two-step objective penalizes both the mean PDE violation after refinement and its variability under the model's own renoise--denoise kernel. The second term is important: a sample may have small residual at a single point yet become non-physical after a small self-consistency perturbation, which the one-step loss cannot detect.

Moreover, if $\boldsymbol f_\theta$ and $\mathcal R$ are differentiable and $b_{t'}$ is small, then
\begin{equation}
\operatorname{tr}\Big(\operatorname{Cov}_{\mathbf{z}} (\mathcal{R}(\mathbf{Y}_{\mathbf{x},t'}) ) \Big) = b_{t'}^{2} {\big\|\mathbf{J}_{\mathcal{R}}\big (\boldsymbol{f}_\theta(a_{t'}\mathbf{x},t') \big)\, \mathbf{D}_{\mathbf{x}} \boldsymbol{f}_{\theta} ( a_{t'}\mathbf{x},t' ) \big\|}_{F}^{2} + o(b_{t'}^{2}).
\end{equation}
So the two-step loss adds a first-order stability penalty on refinement directions that amplify PDE violation. Geometrically, near a regular physical point $\mathbf y \in \mathcal S := \{\mathbf u : \mathcal R(\mathbf u)=0 \}$, the tangent space is $T_{\mathbf y}\mathcal S = \ker \mathbf J_{\mathcal R}(\mathbf y)$; hence this term measures how much the local renoise--denoise dynamics moves away from the tangent space of the physical solution manifold. In this sense, the two-step residual favors not only low PDE error, but also local physical stability of the refinement trajectory, which is especially desirable for few-step consistency sampling.

In principle, one could impose analogous penalties at more intermediate steps, but this would substantially increase training cost. Empirically, we found that the 2-step loss gives a good tradeoff between efficiency and performance.

\end{newsection}

\subsection{Inference and Forward Problem Solving}
\label{subsec:inference}

At inference time, our framework supports two distinct modes of operation. When sampling random solution--coefficient pairs from the learned joint distribution, we directly apply the standard consistency model sampling procedure, enabling fast generation via single-step denoising updates.

In contrast, solving the forward problem—where the coefficient field $\mathbf{a}$ is given and the solution $\mathbf{u}$ is sought—requires a conditional generation mechanism. In this setting, naively applying unconditional sampling is insufficient, as the model must respect the prescribed coefficients while satisfying the governing PDE. We therefore reformulate the forward problem as a \textit{zero-shot inpainting} task.

We introduce an iterative inference procedure that treats the consistency model as a learned solver. Leveraging the zero-shot inpainting strategy proposed in the original Consistency Models framework~\cite{song2023consistency}, we strictly enforce the known coefficient channels $\mathbf{a}_{\text{obs}}$ via a hard projection operator $\mathcal{P}$ at each denoising step. This guides the generative trajectory to evolve on the manifold defined by the specific physical instance. Formally, for a generated state $\hat{\mathbf{x}}$ and a binary mask $\mathbf{M}$ (where $\mathbf{M}=1$ for coefficient channels and $\mathbf{0}$ otherwise), the projection is defined as:
\begin{equation}
    \mathcal{P}(\hat{\mathbf{x}}, \mathbf{x}_{\text{obs}}) = \mathbf{M} \odot \mathbf{x}_{\text{obs}} + (\mathbf{1} - \mathbf{M}) \odot \hat{\mathbf{x}}.
\end{equation}
Importantly, this procedure does not involve explicit numerical optimization or gradient-based PDE residual evaluation at inference time. Instead, the trained model implicitly encodes the coupling between $\mathbf{a}$ and $\mathbf{u}$. The iterative "predict-project-renoise" updates act to align the generated solution with the imposed coefficients, leveraging the robustness induced by our two-step training objective to correct approximation errors.

Algorithmic details of this projection-based forward solver are summarized in Appendix~\ref{app:algo}.

\section{Experiments}

We evaluate the efficacy of our framework on several canonical steady-state PDE benchmarks \modtext{(Darcy flow, Poisson, and Helmholtz equations), as well as the time-dependent Navier-Stokes equations.} We adopt the pre-computed datasets released by the DiffusionPDE authors~\citep{diffusionpde}. These datasets were generated using high-fidelity finite element method (FEM) solvers on a $128 \times 128$ spatial grids, providing $50,000$ paired samples for each equation family.
\begin{newsection}Unless otherwise specified, we utilize the $L^2$ norm ($p=2$) for the PDE residual penalty across all evaluations.\end{newsection}


\subsection{PDE Problem Settings}
\label{subsec:pde_settings}

\begin{modsection}
    We consider a suite of standard elliptic PDEs alongside the time-dependent Navier-Stokes equations. Full mathematical definitions are provided in Appendix~\ref{app:pde}.
\end{modsection}

\begin{modsection}
\subsection{Evaluation Metric: Relative $H^1$ Error}
\label{subsubsec:h1_metric}
\end{modsection}
For forward solution prediction in elliptic PDE benchmarks, prior work often relies solely on the relative $L^2$ error to quantify solution accuracy. However, this metric can be insufficient for physics-informed assessment, as it measures pointwise proximity but is insensitive to high-frequency oscillations or gradient mismatches. As a result, a candidate solution may achieve a low $L^2$ error while exhibiting non-physical roughness that violates the governing differential equation.

To more rigorously assess physical validity in forward solution prediction, we adopt the relative $H^1$ error. This metric accounts for both solution values and their spatial derivatives, ensuring that predicted fields are not only accurate but also smooth and consistent with the underlying physics. Different metrics are employed for conditional inference and unconditional sampling, and are introduced in the corresponding experimental sections.

\paragraph{Definition.}
For a predicted solution $\hat u$ and ground-truth solution $u$ defined on a domain $\Omega$, the squared $H^1$ norm is defined as
\begin{equation}
    \| u \|_{H^1(\Omega)}^2 \triangleq \| u \|_{L^2(\Omega)}^2 + \| \nabla u \|_{L^2(\Omega)}^2
    = \int_\Omega |u|^2 \, d\mathbf{x} + \int_\Omega |\nabla u|^2 \, d\mathbf{x}.
\end{equation}
The relative $H^1$ error is then given by
\begin{equation}
    \mathcal{E}_{H^1} = \frac{\| \hat u - u \|_{H^1(\Omega)}}{\| u \|_{H^1(\Omega)}}.
\end{equation}

\paragraph{Implementation.}
Since the solutions are defined on a discrete grid with spacing $h$, spatial derivatives are approximated using second-order central differences at interior points and first-order differences at the boundaries. The discrete $H^1$ norm used in our evaluation is
\begin{equation}
    \| e \|_{H^1}^2 \approx \sum_{i,j} \left( |e_{i,j}|^2 + |\delta_x e_{i,j}|^2 + |\delta_y e_{i,j}|^2 \right) h^2,
\end{equation}
where $e = \hat u - u$ denotes the error field, and $\delta_x, \delta_y$ are discrete central difference operators (e.g., $\delta_x e_{i,j} \approx \frac{e_{i+1,j} - e_{i-1,j}}{2h}$).

\subsection{Implementation and Baselines}
\label{subsec:implementation}

To assess the impact of our physics-informed fine-tuning, we compare the proposed method against its pre-trained base model and established diffusion-based solvers.

\paragraph{Network Architecture.}
To ensure a fair comparison with the baseline, we adopt the U-Net backbone used in DiffusionPDE~\cite{diffusionpde}, which is based on the DDPM++ architecture~\cite{song2021scorebased}. We modify only the parameterization and normalization components relevant to consistency training. Specifically, while DiffusionPDE employs the EDM formulation~\cite{Karras2022edm} with diffusion-specific pre-conditioning, we remove these scaling factors and operate directly on the underlying U-Net trunk. We then apply the TrigFlow parameterization described in Sec.~\ref{subsec:trigflow_param}.

To improve training stability, we further adjust the normalization layers. As observed in the sCM framework~\cite{lu2025simplifying}, the standard AdaGN layer can lead to divergence in consistency models; accordingly, we replace it with a modified Adaptive Double Normalization that incorporates pixel normalization~\citep{karras2018progressive}.

\paragraph{Training Protocol.}

Training is conducted in three sequential phases. We employ the AdamW optimizer~\cite{loshchilov2018decoupled} throughout all stages, with hyperparameters summarized in Appendix~\ref{app:hyperparameter}.
\begin{enumerate}
    \item \textbf{Diffusion Pre-training:}  We first train a score-based diffusion model to approximate the data distribution $p_{\text{data}}(\mathbf{a}, \mathbf{u})$.
    \item \textbf{Stage 1 (Consistency Training):} We initialize the consistency model from the pre-trained diffusion weights and optimize it using the standard sCM objective (Eq.~\ref{eq:scm_loss_stage1}) to enforce self-consistency.
    \item \textbf{Stage 2 (Physics Fine-Tuning):} Starting from the Stage~1 checkpoint, we fine-tune the model using the physics-informed joint Min--Max objective (Eq.~\ref{eq:total_loss}), with the coefficient decoder frozen. The balancing parameters $\bm{\lambda}$ are initialized to $\bm{\lambda}_{\text{init}}$ (Table~\ref{tab:hyperparams}) and optimized jointly with the model weights.
\end{enumerate}

\paragraph{Metric: Number of Function Evaluations (NFE).}
To quantify inference efficiency, we report the Number of Function Evaluations (NFE) rather than raw sampling steps. This metric directly reflects the number of neural network forward passes and is independent of the specific numerical solver. DiffusionPDE employs Heun's second-order solver~\cite{Karras2022edm} together with Diffusion Posterior Sampling (DPS)~\cite{chung2023diffusion}. This predictor--corrector scheme requires two network evaluations per sampling step, leading to an NFE of $2T-1$ for $T$ sampling steps. In contrast, our sCM-PINN supports single-pass generation, requiring only one network evaluation per step ($\mathrm{NFE}=T$).

\subsection{Forward Problem Results}
\label{subsec:forward_results}

Table~\ref{tab:main_results} summarizes the quantitative results for forward PDE solving. All reported metrics are computed as averages over $256$ randomly selected test samples. We evaluate the models under varying computational budgets to examine the trade-off between inference efficiency and physical accuracy.

\paragraph{Accuracy and Stability}
Our physics-informed model (sCM-PINN) consistently outperforms the Stage~1 data-only baseline (sCM) across all three physical systems. For the Darcy Flow benchmark, physics-informed fine-tuning reduces the relative $H^1$ error by approximately 42\% at \modtext{65} steps ($1.05 \times 10^{-1}$ vs.\ $1.81 \times 10^{-1}$), demonstrating that the frozen-decoder strategy successfully injects physical constraints without degrading the learned data distribution. Importantly, sCM-PINN also achieves accuracy comparable to or better than the DiffusionPDE baseline while operating at substantially fewer function evaluations.

\paragraph{Computational Efficiency}
Thanks to the consistency formulation, sCM-PINN converges rapidly to physically accurate solutions. On the Darcy Flow benchmark, sCM-PINN reaches a relative error of $1.10 \times 10^{-1}$ in only \modtext{17} evaluations, whereas DiffusionPDE exhibits a substantially higher error ($3.49 \times 10^{-1}$) even after 127 evaluations. This corresponds to an effective inference speedup of \modtext{nearly 7.5$\times$} in terms of network evaluations. On the Poisson equation, sCM-PINN attains a relative error of $1.81 \times 10^{-1}$ with \modtext{65} evaluations, slightly improving upon DiffusionPDE with 127 evaluations ($1.85 \times 10^{-1}$).

\begin{table}[h]
  \centering
    \caption{Comparison of relative $H^1$ error for the forward problem averaged over 256 test samples. The best results are highlighted in bold. NFE denotes Number of Function Evaluations.}
  \label{tab:main_results}
  \resizebox{\columnwidth}{!}{
  \begin{tabular}{l c c c c c}
    \toprule
    Method          & NFE & Darcy Flow                 & Poisson                    & Helmholtz                  & \newtext{Navier-Stokes} \\
    \midrule
    DiffusionPDE    & 31  & $6.37 \times 10^{-1}$      & $9.99 \times 10^{-1}$      & $1.38 \times 10^{0}$       & \newtext{$1.13 \times 10^{0}$} \\
                    & 63  & $4.18 \times 10^{-1}$      & $5.34 \times 10^{-1}$      & $1.20 \times 10^{0}$       & \newtext{$8.65 \times 10^{-1}$} \\
                    & 127 & $3.49 \times 10^{-1}$      & $1.85 \times 10^{-1}$      & $9.37 \times 10^{-1}$      & \newtext{$3.85 \times 10^{-1}$} \\
    \addlinespace
    sCM             & \modtext{17}  & $2.00 \times 10^{-1}$      & $4.17 \times 10^{-1}$      & $4.31 \times 10^{-1}$      & \newtext{$2.79 \times 10^{-1}$} \\
                    & \modtext{33}  & $1.92 \times 10^{-1}$      & $3.33 \times 10^{-1}$      & $3.34 \times 10^{-1}$      & \newtext{$2.40 \times 10^{-1}$} \\
                    & \modtext{65}  & $1.81 \times 10^{-1}$      & $2.78 \times 10^{-1}$      & $2.82 \times 10^{-1}$      & \newtext{$2.12 \times 10^{-1}$} \\                  
    \addlinespace
    sCM-PINN (Ours) & \modtext{17}  & $1.10 \times 10^{-1}$      & $3.20 \times 10^{-1}$      & $3.76 \times 10^{-1}$      & \newtext{$2.06 \times 10^{-1}$} \\
                    & \modtext{33}  & $1.08 \times 10^{-1}$      & $2.28 \times 10^{-1}$      & $2.87 \times 10^{-1}$      & \newtext{$1.62 \times 10^{-1}$} \\
                    & \modtext{65}  & $\mathbf{1.05 \times 10^{-1}}$ & $\mathbf{1.81 \times 10^{-1}}$ & $\mathbf{2.33 \times 10^{-1}}$ & \newtext{$\mathbf{1.43 \times 10^{-1}}$} \\
                    
    \bottomrule
  \end{tabular}
  }
\end{table}

\paragraph{Wall-Clock Efficiency.}
While the number of function evaluations (NFE) provides a hardware-agnostic measure of sampling complexity, it does not fully capture the computational cost of diffusion-based solvers that rely on gradient-based guidance. In particular, DiffusionPDE employs diffusion posterior sampling (DPS)~\citep{chung2023diffusion}, where each sampling step incurs the additional overhead of backpropagating through the PDE residual operator.

To account for this discrepancy, we additionally report wall-clock inference time (See Appendix~\ref{app:walltime} for details), comparing sCM-PINN with DiffusionPDE for generating a single sample (batch size = 1).
Despite using a similar number of steps (\modtext{65} vs.\ 63), sCM-PINN achieves substantially lower inference time ($1.56$s vs.\ $3.86$s), as each step consists solely of a forward network evaluation, whereas DPS requires repeated gradient computations.
\begin{newsection}
Furthermore, we benchmarked sCM-PINN against a traditional CPU Immersed Interface Method (IIM) solver~\cite{doi:10.1137/1.9780898717464} for Darcy flow. Averaged over 100 samples, the IIM solver takes 1.40s per sample, demonstrating that sCM-PINN reaches comparable practical clock time to a traditional numerical solver, while additionally supporting joint generative modeling.
\end{newsection}

\subsection{Conditional Source Reconstruction Results}

We evaluate the framework's ability to perform \emph{conditional generative inference}, in which the source term $a(\mathbf{x})$ is inferred from partial or complete observations of the solution field $u(\mathbf{x})$. This task corresponds to an ill-posed inverse setting, in which multiple source configurations may be consistent with the same observed solution. We consider Poisson and Helmholtz benchmarks and report quantitative results in Table~\ref{tab:inverse_results}.

\paragraph{Metric Selection (Relative $L^2$ vs.\ $H^1$).}
Unlike the forward problem, we adopt the relative $L^2$ error as the primary metric for conditional inference. This choice reflects the ill-posed nature of inverse PDE settings, which are inherently sensitive to high-frequency perturbations and may admit multiple valid reconstructions. Moreover, while the solution field $u$ is typically smooth due to elliptic regularity, the source term $a$ is not necessarily differentiable and may contain sharp transitions or discontinuities. 

\begin{newsection}
Formally, the $H^1$ norm requires a function and its first weak derivative to be both square-integrable (i.e., both $u \in L^2$ and $\nabla u \in L^2$). Specifically, for the solution $u$ of the Poisson and Helmholtz equations to exist in $H^1$, the source term $a$ only needs to belong to the dual space $H^{-1}$. Even if $a \in L^2$ (which implies $u \in H^2$), the source term $a$ might not belong to $H^1$.
\end{newsection}

As a result, an $H^1$-based metric would disproportionately penalize high-frequency components and overemphasize small-scale artifacts, leading to a misleading assessment of reconstruction quality. We emphasize that our goal is not to recover a unique ground-truth source, but rather to generate physically consistent source realizations conditioned on the observed solution.

\begin{table}[h]
    \centering
    \caption{Comparison of relative $L^2$ error for conditional source inference averaged over $256$ test samples. The best results are highlighted in bold. NFE denotes the Number of Function Evaluations.}
    \label{tab:inverse_results}
    \begin{tabular}{l c c c}
        \toprule
        Method          & NFE & Poisson                    & Helmholtz                  \\
        \midrule
        DiffusionPDE    & 31  & $1.21 \times 10^{0}$       & $1.34 \times 10^{0}$       \\
                        & 63  & $8.91 \times 10^{-1}$      & $1.20 \times 10^{0}$       \\
                        & 127 & $4.56 \times 10^{-1}$      & $9.74 \times 10^{-1}$      \\
        \addlinespace
        sCM-PINN (Ours) & \modtext{17}  & $5.92 \times 10^{-1}$      & $5.72 \times 10^{-1}$      \\
                        & \modtext{33}  & $4.85 \times 10^{-1}$      & $4.58 \times 10^{-1}$      \\
                        & \modtext{65}  & $\bm{3.99 \times 10^{-1}}$ & $\bm{3.81 \times 10^{-1}}$ \\
        \bottomrule
    \end{tabular}
\end{table}

\subsection{Unconditional Sampling}

Beyond forward problem solving and conditional inference, we evaluate the quality of \emph{unconditional} samples generated by each model. In this setting, samples are drawn directly from the learned joint distribution $p(\mathbf{a}, \mathbf{u})$ without conditioning on observations. This experiment assesses whether the generative prior learned by each method captures both the data distribution and the governing physical constraints in a free-form generation regime.

\begin{newsection}
For unconditional sampling, the exact solution is not available, so evaluation must rely on PDE residuals rather than solution errors. In this setting, an $H^1$-type error is not well-defined, and it is more appropriate to measure the $L^2$ norm of the residual to assess physical consistency. 
\end{newsection}

To quantify physical fidelity, we compute the normalized PDE residual $\|\mathcal{R}\|_2 \cdot h^2$, where $\mathcal{R}$ denotes the residual operator and $h$ is the grid spacing. Lower values indicate closer adherence to the governing equations. Since unconditional samples are not constrained by observations, this metric evaluates physical plausibility rather than exact solution accuracy, reflecting the quality of the learned generative prior.

Table~\ref{tab:uncond_results} presents the quantitative comparison. Notably, \textbf{sCM-PINN} achieves the lowest residual errors across all three benchmarks while requiring only \textbf{2 function evaluations (NFE)}. This suggests that physics-informed fine-tuning improves not only conditional accuracy but also aligns the learned generative prior more closely with the underlying physical laws. In contrast, the standard sCM baseline exhibits substantially higher residuals (up to $20\times$ higher on Darcy Flow), indicating that data-driven training alone struggles to enforce physical consistency in unconditional generation. While DiffusionPDE performs robustly, sCM-PINN matches or exceeds its physical fidelity with a $\mathbf{30\times}$ reduction in computational cost.

\begin{figure}[h]
    \centering
    \includegraphics[width=.85\linewidth]{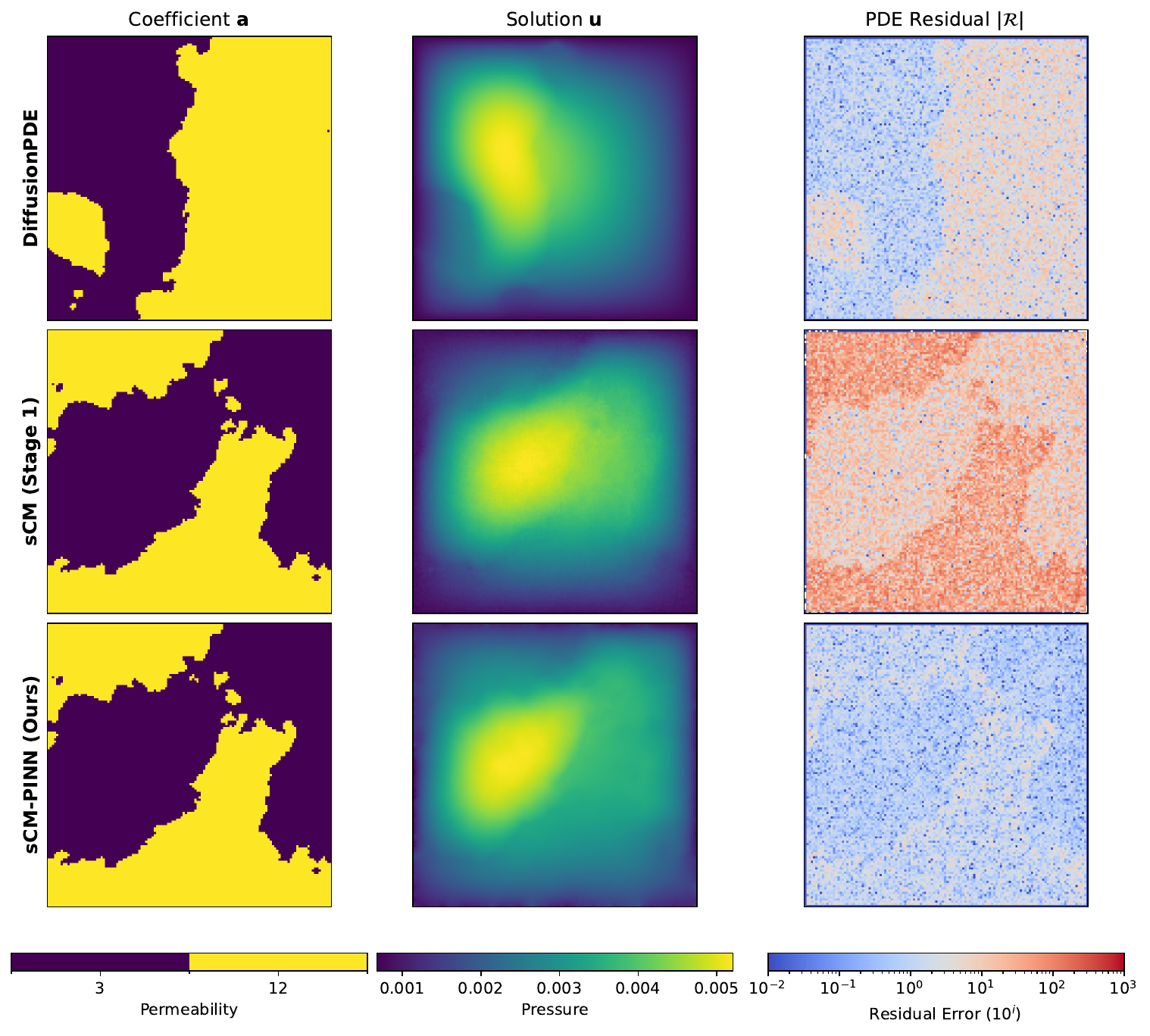}
    \caption{\textbf{Unconditional Sampling Quality (Darcy Flow).} Visual comparison of generated coefficient fields $\mathbf{a}$ (left), solution fields $\mathbf{u}$ (middle), and absolute PDE residuals $|\mathcal{R}|$ (right). The baseline sCM (middle row) exhibits visible artifacts, whereas sCM-PINN (bottom row) produces sharper coefficient interfaces and solution fields with lower and more uniform residuals, comparable to those obtained by the 63-step DiffusionPDE baseline (top row) while requiring only 2 function evaluations. Note the reduced residual variance in the sCM-PINN samples relative to the diffusion baseline.}
    \label{fig:uncond}
\end{figure}

We visualize unconditional samples for Darcy Flow in Figure~\ref{fig:uncond}. Qualitatively, sCM-PINN produces sharp, coherent coefficient interfaces and smooth solution fields that closely resemble the reference data. The residual maps further reveal that DiffusionPDE tends to produce localized high-error regions in high-permeability zones ($a=12$), whereas sCM-PINN maintains more uniform physical consistency across the domain.

\begin{table}[h]
    \centering
    \caption{Comparison of unconditional sampling quality. We report the normalized PDE residual ($\|\mathcal{R}\|_2 \cdot h^2$), where lower values indicate better physical fidelity.}
    \label{tab:uncond_results}
    \resizebox{\columnwidth}{!}{
    \begin{tabular}{l c c c c c}
        \toprule
        Method          & NFE & Darcy Flow                 & Poisson                    & Helmholtz                  & \newtext{Navier-Stokes} \\
        \midrule
        DiffusionPDE    & 63  & $3.22 \times 10^{-2}$      & $1.18 \times 10^{-2}$      & $1.99 \times 10^{-2}$      & \newtext{$1.92 \times 10^{-2}$} \\
        \addlinespace
        sCM             & 2   & $4.44 \times 10^{-1}$      & $4.05 \times 10^{-2}$      & $4.67 \times 10^{-2}$      & \newtext{$2.12 \times 10^{-2}$} \\
        \addlinespace
        sCM-PINN (Ours) & 2   & $\bm{2.00 \times 10^{-2}}$ & $\bm{1.13 \times 10^{-2}}$ & $\bm{1.11 \times 10^{-2}}$ & \newtext{$\mathbf{7.52 \times 10^{-3}}$} \\
        \bottomrule
    \end{tabular}
    }
\end{table}

\subsection{Ablation Study: Effect of Freezing the Coefficient Decoder}
\label{subsec:ablation_decoder}

In this section, we investigate the role of freezing the coefficient decoder during Stage~2 physics-informed fine-tuning. We compare our proposed approach against a baseline where the entire model is trained jointly without freezing any components. For this joint training baseline, we modify the Simplified Consistency Term (Eq.~\ref{eq:scm_loss_stage2}) in the total loss to explicitly supervise the coefficient channel. Specifically, we replace the component-wise masked loss with a full-state $L_2$ loss over both coefficient and solution channels:
\begin{equation*}
  \ell_{\text{sCM-L2-joint}}(\theta; \mathbf{x}_t, t) = \left\| F_\theta\!\left(\tfrac{\mathbf{x}_t}{\sigma_d}, t\right) - \mathbf{y}_t \right\|_2^2.
\end{equation*}

\paragraph{Coefficient Mode Collapse}
In the Darcy Flow problem, the permeability field $\mathbf{a}(\mathbf{x})$ takes discrete values in $\{3, 12\}$ with approximately equal probability. When we allow the coefficient decoder to update during physics fine-tuning, the model rapidly collapses to a unimodal distribution concentrated at $a \approx 12$. We attribute this behavior to the observation that higher permeability values yield smoother pressure fields $\mathbf{u}$ for a fixed forcing term. Consequently, the optimizer minimizes the PDE residual by shifting the coefficient distribution toward these simpler configurations, effectively ignoring the target data distribution learned in Stage~1.

\begin{figure}[ht]
    \centering
    \includegraphics[width=.9\linewidth]{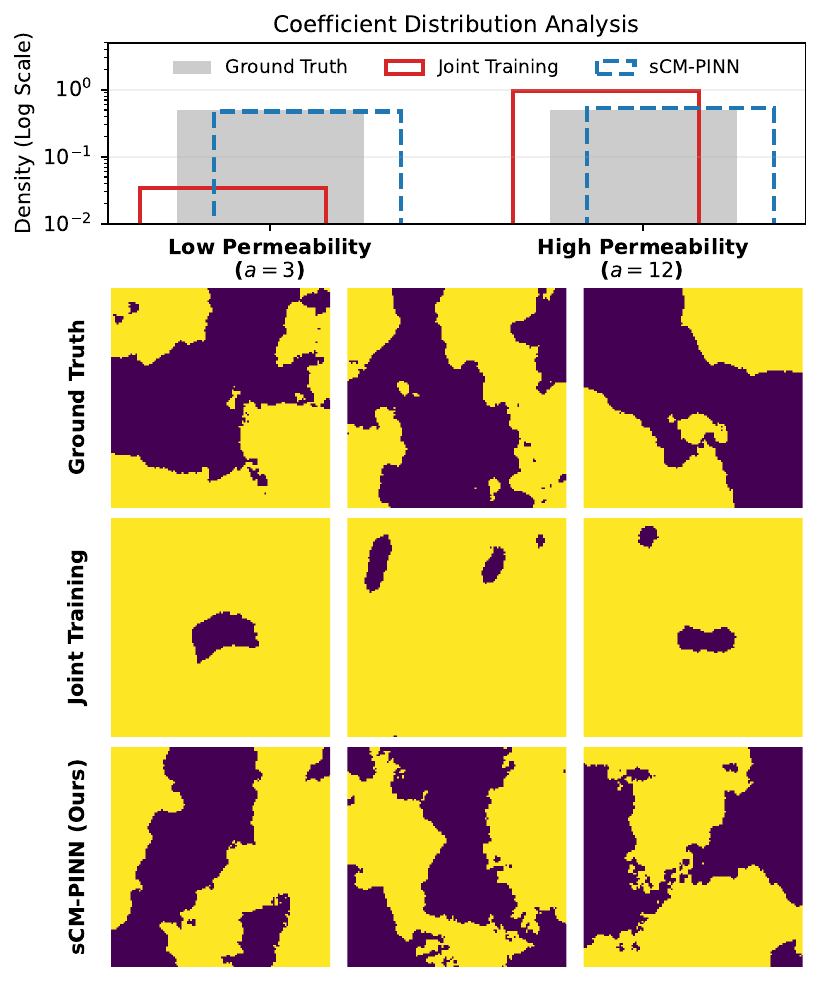}
    \caption{\textbf{Analysis of Coefficient Mode Collapse.} 
    \textbf{(Top)} Pixel-wise histogram showing the bimodal ground-truth distribution (gray). Joint training (red) collapses the distribution to a single mode at $a \approx 12$.
    \textbf{(Bottom)} Representative spatial samples. The joint-training baseline (middle row) produces only high-permeability features, whereas the frozen decoder strategy (bottom row) recovers the full coefficient structure.}
    \label{fig:ablation_mode_collapse}
\end{figure}

\paragraph{Quantitative Analysis}
Figure~\ref{fig:ablation_mode_collapse} compares the pixel-wise histograms and spatial coefficient maps of the generated fields. We generate 128 samples from each model using two-step sampling, visualizing both pixel-wise value distributions and representative spatial coefficient maps. The joint training baseline (red) exhibits severe mode collapse, failing to capture the low-permeability mode properly. In contrast, our frozen decoder approach (blue) preserves the ground-truth statistics (gray) while still minimizing the PDE residual. These results indicate that disentangling input generation (Stage~1, fixed) from solution optimization (Stage~2, active) is critical for preserving the learned coefficient distribution while enforcing physical consistency.

\begin{newsection}
\subsection{Extended Experimental Results}
\label{subsec:additional_experiments}
To broaden the scope of our evaluation, we conducted extended experiments (Appendix~\ref{app:additional_experiments}). 

\textbf{Deterministic Operators:} We compare against FNO~\cite{li2021fourier} and DeepONet~\cite{osti_2281727} for forward problems. While FNO excels on Poisson and Helmholtz, it is a strictly supervised model. Conversely, sCM-PINN natively supports stochastic modeling and outperforms FNO on Darcy Flow (Table~\ref{tab:operator_baselines}). 

\textbf{Distributional Fidelity:} We benchmark against concurrent physics-informed models (PBFM, PIDM) on their specific Darcy Flow dataset. Evaluated via Wasserstein distance (WD) and Jensen-Shannon divergence (JS), sCM-PINN maintains distributional quality orders of magnitude better than these baselines at just 2 NFE (Table~\ref{tab:generative_metrics}).

\textbf{Navier-Stokes System:} Building upon the forward and unconditional results presented earlier, we provide extended evaluations for this time-dependent system in Appendix~\ref{subsec:app_navier}. For the inverse problem (Table~\ref{tab:navier_inverse}), sCM-PINN achieves a lower $H^1$ error than DiffusionPDE in fewer steps. For unconditional generation (Table~\ref{tab:navier_wd_js}), we analyze the computational trade-off, showing sCM-PINN achieves the lowest PDE residual at just 2 NFE, and recovers highly competitive distributional fidelity (WD and JS) when scaled to 17 NFE.

\textbf{Sensitivity and Training Cost:} Our method transfers robustly without exhaustive tuning. Transitioning to Stage 2 at varying checkpoints yields stable metrics, proving precise scheduling is unnecessary (Table~\ref{tab:stage_transition}). Total training requires $\sim$17 hours on an RTX 4090.
\end{newsection}

\section{Conclusion}

In this work, we introduced \emph{sCM-PINN}, a physics-informed consistency modeling framework for efficient PDE solving and physically grounded generative modeling. We identified key stability challenges that arise when enforcing physical constraints in consistency training, where naive joint optimization can lead to degenerate solutions and coefficient mode collapse. To address this issue, we proposed a structure-preserving training strategy based on a two-stage protocol that couples a frozen coefficient decoder with a physics-informed residual objective. This design stabilizes optimization, enabling the model to respect physical laws while preserving the learned generative prior.

Across Darcy Flow, Poisson, and Helmholtz benchmarks, sCM-PINN consistently achieves strong physical accuracy with substantially fewer function evaluations than diffusion-based PDE solvers. In particular, sCM-PINN attains the lowest relative $H^1$ error among the evaluated methods while requiring only a small number of model evaluations at inference. Ablation studies further demonstrate that freezing the coefficient decoder during physics fine-tuning is critical for preventing mode collapse and maintaining the diversity of the learned coefficient distribution. Moreover, unconditional sampling experiments show that sCM-PINN produces physically plausible samples with only two function evaluations, whereas diffusion-based baselines require orders of magnitude more computational effort. By bridging the efficiency of consistency models with the rigor of physics-informed learning, sCM-PINN offers a promising direction toward real-time scientific simulation and generative modeling of physical systems. Future work will explore extensions to more complex PDEs, time-dependent dynamics, and higher-dimensional settings.

\begin{newsection}
\section{Limitations and Ethical Considerations}
\label{sec:limitations}
Our multi-phase training pipeline introduces additional implementation overhead, and the continuous-time formulation requires Jacobian-Vector Products (JVPs), increasing memory cost. As our experiments use only synthetic open-source PDE data, we foresee no ethical concerns.
\end{newsection}

\begin{acks}
T.-S. Lin acknowledges the supports by National Science and Technology Council, Taiwan, under research grants 111-2628-M-A49-008-MY4 and 114-2124-M-390-001. M.-C. Lai acknowledges the support by National Science and Technology Council, Taiwan, under research grant 113-2115-M-A49-014-MY3.
\end{acks}

\section*{GenAI Disclosure}

The authors acknowledge the use of large language models to polish and edit the written text of the manuscript.

\bibliographystyle{ACM-Reference-Format}
\balance
\bibliography{ref}

\appendix

\section{Algorithm Details}
\label{app:algo}

We provide the complete pseudocode for the projection-based forward solver detailed in Algorithm~\ref{alg:forward_inference}.

\begin{algorithm}[H]
\caption{Measurement-Constrained Consistency Sampling (TrigFlow)}
\label{alg:forward_inference}
\begin{algorithmic}[1]
\REQUIRE Consistency Model $\bm{f}_\theta(\mathbf{x}, t)$, Observation $\mathbf{x}_{\text{obs}}$, Mask $\mathbf{M}$.
\REQUIRE Time schedule $t_0 > t_1 > \dots > t_N$ (where $t_0 \approx \pi/2, t_N \approx 0$).
\REQUIRE Data deviation $\sigma_d$.
\STATE \textbf{Initialize:} Sample noise $\mathbf{z} \sim \mathcal{N}(\mathbf{0}, \mathbf{I})$.
\STATE Set initial noisy state $\mathbf{x}_{t_0} \leftarrow \sigma_d \mathbf{z}$ (assuming $t_0 \approx \pi/2$).
\STATE \textbf{Initial Estimation:}
\STATE $\hat{\mathbf{x}} \leftarrow \bm{f}_\theta(\mathbf{x}_{t_0}, t_0)$ \COMMENT{Predict $\mathbf{x}_0$ using Eq.~\ref{eq:trigflow}}
\STATE $\hat{\mathbf{x}} \leftarrow \mathbf{M} \odot \mathbf{x}_{\text{obs}} + (\mathbf{1}-\mathbf{M}) \odot \hat{\mathbf{x}}$ \COMMENT{Apply measurement constraint}
\FOR{$n = 1$ \TO $N$}
    \STATE Sample fresh noise $\mathbf{z} \sim \mathcal{N}(\mathbf{0}, \mathbf{I})$.
    \STATE \textbf{Re-noising (TrigFlow):}
    \STATE $\mathbf{x}_{t_n} \leftarrow \cos(t_n)\hat{\mathbf{x}} + \sin(t_n)\sigma_d \mathbf{z}$ \COMMENT{Back-project to manifold at $t_n$}
    \STATE \textbf{Consistency Update:}
    \STATE $\hat{\mathbf{x}} \leftarrow \bm{f}_\theta(\mathbf{x}_{t_n}, t_n)$ \COMMENT{Jump to data origin}
    \STATE \textbf{Projection:}
    \STATE $\hat{\mathbf{x}} \leftarrow \mathbf{M} \odot \mathbf{x}_{\text{obs}} + (\mathbf{1}-\mathbf{M}) \odot \hat{\mathbf{x}}$ \COMMENT{Enforce measurements}
\ENDFOR
\RETURN $\hat{\mathbf{x}}$
\end{algorithmic}
\end{algorithm}

\section{PDE Problem Details}
\label{app:pde}

\paragraph{Darcy Flow.}
Darcy flow describes fluid transport through porous media and serves as a standard benchmark in hydrogeology and reservoir modeling. We consider the steady-state 2D Darcy flow equation on the unit square $\Omega=(0, 1)^2$:
\begin{equation}
    \begin{aligned}
        -\nabla \cdot \big(a(\mathbf{x}) \nabla u(\mathbf{x})\big) &= f(\mathbf{x}), \quad && \mathbf{x} \in \Omega,
    \end{aligned}
\end{equation}
where $u(\mathbf{x})$ denotes the fluid pressure and $a(\mathbf{x})$ is the permeability field. We evaluate on two commonly used Darcy-flow benchmarks that differ in their coefficient fields, forcing terms, and boundary conditions.

\paragraph{DiffusionPDE Benchmark.}
The DiffusionPDE benchmark is discretized on a $128 \times 128$ spatial grid over the unit square domain $\Omega=(0,1)^2$, with homogeneous Dirichlet boundary conditions,
\begin{equation}
    u(\mathbf{x}) = 0,
    \qquad
    \mathbf{x}\in\partial\Omega.
\end{equation}
The source term is fixed as $f(\mathbf{x})=1$. The permeability field is a piecewise-constant medium taking values in $\{3,12\}$, representing a composite material with distinct permeability regions.

\paragraph{PBFM/PIDM Benchmark.}
We also consider a second Darcy flow benchmark adopted in recent studies on physics-informed generative modeling~\cite{bastek2025physicsinformed,baldan2026physics}. This benchmark is discretized on a $64\times64$ spatial grid and imposes homogeneous Neumann boundary conditions together with a mean-zero pressure constraint:
\begin{equation}
    \begin{aligned}
        \nabla u(\mathbf{x})\cdot \hat n(\mathbf{x}) &= 0,
        \qquad && \mathbf{x}\in\partial\Omega,\\
        \int_\Omega u(\mathbf{x})\,d\mathbf{x} &= 0.
    \end{aligned}
\end{equation}
The permeability field is modeled as a continuous log-normal Gaussian random field,
$
a(\mathbf{x})=\exp(G(\mathbf{x})),
$
where $G(\mathbf{x})$ is a Gaussian random field. The source term is given by
\begin{equation}
    f(\mathbf{x})=
    \begin{cases}
        r, & \text{if } \left|x_i-\frac{w}{2}\right|\le \frac{w}{2},\\
        -r, & \text{if } \left|x_i-1+\frac{w}{2}\right|\le \frac{w}{2},\\
        0, & \text{otherwise},
    \end{cases}
\end{equation}
with $r=10$ and $w=0.125$.

\paragraph{Poisson Equation.}
As a fundamental elliptic operator, the Poisson equation models potential fields arising in electrostatics, gravitation, and steady-state heat conduction. We solve
\begin{equation}
    \begin{aligned}
        \nabla^2 u(\mathbf{x}) &= a(\mathbf{x}), \quad && \mathbf{x} \in \Omega, \\
        u(\mathbf{x}) &= 0, && \mathbf{x} \in \partial \Omega.
    \end{aligned}
\end{equation}
In this setting, $a(\mathbf{x})$ acts as a spatially varying source term, while $u(\mathbf{x})$ represents the resulting potential field. The source terms are synthesized using Gaussian random fields (GRFs), yielding a diverse collection of smooth input structures.

\paragraph{Inhomogeneous Helmholtz Equation.}
To assess the model’s ability to handle wave-like phenomena, we consider the inhomogeneous Helmholtz equation, which augments the Poisson operator with a reaction term:
\begin{equation}
    \begin{aligned}
        \nabla^2 u(\mathbf{x}) + k^2 u(\mathbf{x}) &= a(\mathbf{x}), \quad && \mathbf{x} \in \Omega, \\
        u(\mathbf{x}) &= 0, && \mathbf{x} \in \partial \Omega.
    \end{aligned}
\end{equation}
We fix the wavenumber at $k = 1$. As in the Poisson case, $a(\mathbf{x})$ denotes the source distribution. The addition of the linear reaction term $k^2 u(\mathbf{x})$ introduces oscillatory behavior in the solution manifold, posing a qualitatively different challenge from the purely diffusive Darcy and Poisson systems.

\begin{newsection}
\paragraph{Non-bounded Navier-Stokes Equation.}
We consider the incompressible Navier-Stokes equation for the vorticity.
\begin{equation}
    \begin{aligned}
        \partial_t w(\mathbf{x}, t) + v(\mathbf{x}, t) \cdot \nabla w(\mathbf{x}, t) &= \nu\Delta w(\mathbf{x}, t) + q(\mathbf{x}), \quad && \mathbf{x} \in \Omega, t \in (0,T] \\
        \nabla \cdot v(\mathbf{x}, t) &= 0, \quad && \mathbf{x} \in \Omega, t \in [0,T] 
    \end{aligned}
\end{equation}
Here $w=\nabla\times v$ is the vorticity, $v(\mathbf{x}, t)$ is the velocity at $\mathbf{x}$ at time $t$, and  $q(\mathbf{x})$ is a force field. The viscosity coefficient is set to $\nu=10^{-3}$, which corresponds to the Reynolds number $Re=\frac{1}{\nu}=1000$. 
\end{newsection}

\section{Training Hyperparameter Settings}
\label{app:hyperparameter}

We provide the training hyperparameters for our training protocol, which is summarized in Table~\ref{tab:hyperparams}.

\begin{table}[ht]
\centering
\small
\setlength{\tabcolsep}{2.5pt} 
\caption{Training hyperparameters. \textbf{LR}: Learning Rate, \textbf{BS}: Batch Size. Epochs are listed for (Darcy / Poisson / Helmholtz\newtext{ / Navier-Stokes}). $\bm{\lambda}_{\text{init}}$ denotes the initialization of the learnable loss masks $\{\lambda_0, \lambda_1, \lambda_2\}$ in Stage 2.}
\label{tab:hyperparams}
\begin{tabular}{l c c c c}
\toprule
\textbf{Stage} & \textbf{LR} & \textbf{BS} & $\bm{\lambda}_{\text{init}}$ & \textbf{Epochs (D/P/H\newtext{/N})} \\
\midrule
Pre-train & $1\text{e-}3$ & 16 & -- & 10 / 10 / 10 \newtext{/10}\\
Stage 1 & $1\text{e-}3$ & 2 & -- & 8 / 8 / 8 \newtext{/8}\\
Stage 2 & $1\text{e-}4$ & 2 & $\{10, -10, -10\}$ & 1 / 2 / 2 \newtext{/2}\\
\bottomrule
\end{tabular}
\end{table}

\begin{modsection}
\section{Hardware and Runtime Analysis}
\label{app:walltime}

All inference experiments were conducted on a workstation equipped with an Intel i7-13700 CPU and an NVIDIA RTX 4090 GPU with 24GB of VRAM.

Table~\ref{tab:walltime} compares the wall-clock time required to generate a single Darcy flow solution. First, we compare against the generative baseline. sCM-PINN significantly outperforms DiffusionPDE (1.56s vs. 3.86s) by eliminating the computational overhead of the iterative gradient-based guidance used in DiffusionPDE.
\end{modsection}

\begin{newsection}
Furthermore, to demonstrate practical utility, we benchmarked our sCM-PINN against a traditional CPU Immersed Interface Method (IIM) solver~\cite{doi:10.1137/1.9780898717464} for Darcy flow. Averaged over 100 samples, the IIM solver takes 1.40 seconds per sample. At our maximum forward-problem setting (65 steps, batch size 1), sCM-PINN achieves comparable wall-clock time at 1.56 seconds. We note that since the IIM runs on the CPU and sCM-PINN runs on the GPU, the precise timing ratio will vary with different hardware configurations.
\end{newsection}
\begin{table}[ht]
  \begin{modsection}
  \centering
  \caption{Wall-clock sampling time (in seconds) for generating a single Darcy flow sample.}
  \label{tab:walltime}

  \renewcommand{\arraystretch}{0.8}
  \begin{tabular}{llcc}
    \toprule
    Method & Hardware & NFE & Time (s) \\
    \midrule
    IIM (Traditional Solver) & CPU (i7-13700) & -- & 1.40 \\
    sCM-PINN (Ours)          & GPU (RTX 4090) & 65 & 1.56 \\
    DiffusionPDE             & GPU (RTX 4090) & 63 & 3.86 \\
    \bottomrule
  \end{tabular}
  \end{modsection}
\end{table}

\begin{newsection}
\section{Extended Experimental Results}
\label{app:additional_experiments}

This section provides the comprehensive quantitative tables supporting the additional evaluations discussed in Sec.~\ref{subsec:additional_experiments}.
\end{newsection}

\begin{newsection}
\subsection{Comparison to Deterministic Neural Operators}
To show that our generative approach is competitive in accuracy for forward problems, we compared sCM-PINN against deterministic neural operator baselines, specifically FNO and DeepONet. As shown in Table~\ref{tab:operator_baselines}, sCM-PINN (at NFE=65) achieves competitive forward-solving relative $L^2$ errors against deterministic baselines and DiffusionPDE, outperforming FNO on the Darcy Flow benchmark. 

While FNO excels on Poisson/Helmholtz, it is a strictly supervised, deterministic solver tailored solely for forward tasks. This deterministic nature makes FNO more restrictive for uncertainty quantification, inverse problems, and learning from partial or scarce observations. Conversely, sCM-PINN is a unified generative model that natively supports stochastic modeling, unconditional sampling, forward solving, and source reconstruction.
\end{newsection}
\begin{table}[ht]
  \begin{newsection}
  \centering
  \caption{Relative $L^2$ error comparison against deterministic neural operators.}
  \label{tab:operator_baselines}

  \renewcommand{\arraystretch}{0.8}
  \begin{tabular}{lccc}
    \toprule
    Method & Darcy Flow & Poisson & Helmholtz \\
    \midrule
    DeepONet & $1.23 \times 10^{-1}$ & $1.43 \times 10^{-1}$ & $1.78 \times 10^{-1}$ \\
    FNO & $5.30 \times 10^{-2}$ & $\mathbf{8.20 \times 10^{-2}}$ & $\mathbf{1.11 \times 10^{-1}}$ \\
    DiffusionPDE & $7.31 \times 10^{-1}$ & $2.44 \times 10^{-1}$ & $1.02 \times 10^{0}$ \\
    sCM-PINN & $\mathbf{2.80 \times 10^{-2}}$ & $8.91 \times 10^{-2}$ & $1.36 \times 10^{-1}$ \\
    \bottomrule
  \end{tabular}
  \end{newsection}
\end{table}

\begin{newsection}
\subsection{Distributional Fidelity and Concurrent Baselines}

To demonstrate that sCM-PINN ensures distributional fidelity alongside physical accuracy, we benchmark against concurrent physics-informed generative models (PBFM and PIDM). For this evaluation, we utilize the specific Darcy Flow dataset introduced by PBFM and PIDM, which is distinct from the primary Darcy Flow dataset analyzed in the main paper. To ensure a strictly fair comparison and to address reporting discrepancies regarding generative quality, we evaluated all models using PBFM’s exact residual implementation alongside our distributional metrics.

We evaluate distributional fidelity using the Wasserstein distance (WD) and Jensen-Shannon divergence (JS), while physical fidelity is measured via the normalized PDE residual $\|\mathcal{R}\|_2 \cdot h^2$. During training, sCM-PINN utilized the exact same network hyperparameters from our Helmholtz and Poisson setups, demonstrating its out-of-the-box robustness. Additionally, for the physics-informed fine-tuning on this specific dataset, we set the residual loss norm to $p=1$ (as defined in Eq.~\ref{eq:total_loss}). 

As shown in Table~\ref{tab:generative_metrics}, sCM-PINN drastically reduces PDE residuals while maintaining WD ($1.93\times 10^{-2}$) and JS ($7.23\times 10^{-5}$) orders of magnitude better than PBFM and PIDM at just 2 NFE (vs. 20-100 NFE).
\end{newsection}
\begin{table}[ht]
  \begin{newsection}
  \centering
  \caption{Distributional metrics and physical fidelity on the Darcy Flow dataset.}
  \label{tab:generative_metrics}

\renewcommand{\arraystretch}{0.8}
  \begin{tabular}{lcccc}
    \toprule
    Model & NFE & $\|\mathcal{R}\|_2 \cdot h^2$ & WD & JS \\
    \midrule
    sCM & 2 & $4.26 \times 10^{-3}$ & $\mathbf{1.85 \times 10^{-2}}$ & $7.56 \times 10^{-5}$ \\
    sCM-PINN & 2 & $2.39 \times 10^{-3}$ & $1.93 \times 10^{-2}$ & $\mathbf{7.23 \times 10^{-5}}$ \\
    PBFM & 20 & $\mathbf{2.11 \times 10^{-4}}$ & $2.70 \times 10^{-1}$ & $1.72 \times 10^{-2}$ \\
    PIDM & 100 & $2.98 \times 10^{-4}$ & $9.83 \times 10^{-2}$ & $1.33 \times 10^{-2}$ \\
    \bottomrule
  \end{tabular}
  \end{newsection}
\end{table}

\begin{newsection}
\subsection{Extended Evaluation on the Navier-Stokes System}
\label{subsec:app_navier}

To adapt our generative formulation to the Navier-Stokes system, we map the initial vorticity state $w_0$ to the coefficient channel ($\mathbf{a} = w_0$) and the final state $w_T$ to the solution channel ($\mathbf{u} = w_T$). By jointly modeling the distribution of $(w_0, w_T)$, the frozen-decoder strategy generalizes effectively, yielding a unified inference framework capable of solving both the forward problem (predicting $w_T$ given $w_0$) and the inverse problem (reconstructing $w_0$ from $w_T$). 

Tables \ref{tab:navier_inverse} and \ref{tab:navier_wd_js}\footnote{\textit{Distributional metrics differ from our \href{https://openreview.net/forum?id=80TYrhuL5s&noteId=42AtmXW9FU}{OpenReview rebuttal} due to a corrected evaluation implementation.}} summarize our findings. For the inverse problem, sCM-PINN outperforms DiffusionPDE's $H^1$ error in fewer steps. In the unconditional generative setting, sCM-PINN achieves the lowest overall PDE residual at just 2 NFE. Furthermore, when scaled to 17 NFE, our model surpasses the 63-step DiffusionPDE baseline across all evaluated metrics (PDE residual, WD, and JS), while achieving distributional fidelity (WD/JS) highly comparable to the base sCM.

\begin{table}[ht]
  \begin{newsection}
  \centering
  
  \caption{Inverse problem results (reconstructing initial state $w_0$) on the Navier-Stokes system. We report the relative $H^1$ error.}
  \label{tab:navier_inverse}

\renewcommand{\arraystretch}{0.8}  
  \begin{tabular}{lcc}
    \toprule
    Method & NFE & Inverse $H^1$ \\
    \midrule
    DiffusionPDE    & 31  & $1.33 \times 10^{0}$ \\
        & 63  & $1.17 \times 10^{0}$ \\
        & 127 & $9.17 \times 10^{-1}$ \\
    \addlinespace
    sCM-PINN (Ours) & 17  & $8.17 \times 10^{-1}$ \\
     & 33  & $7.24 \times 10^{-1}$ \\
     & 65  & $\mathbf{6.68 \times 10^{-1}}$ \\
    \bottomrule
  \end{tabular}
  \end{newsection}
\end{table}

\begin{table}[ht]
\begin{newsection}
  \centering
  \caption{Distributional metrics and PDE residuals (unconditional sampling) on the Navier-Stokes system across varying computational budgets.}
  \label{tab:navier_wd_js}

  \begin{tabular}{lcccc}
    \toprule
    Method & NFE & $\|\mathcal{R}\|_2 \cdot h^2$ & WD & JS \\
    \midrule
    DiffusionPDE    & 63 & $1.92 \times 10^{-2}$ & $4.45 \times 10^{-3}$ & $3.45 \times 10^{-4}$ \\
    \addlinespace
    sCM             & 2  & $2.12 \times 10^{-2}$ & $2.40 \times 10^{-2}$ & $2.97 \times 10^{-3}$ \\
                    & 17 & $1.93 \times 10^{-2}$ & $\mathbf{2.08 \times 10^{-3}}$ & $\mathbf{1.23 \times 10^{-4}}$ \\
    \addlinespace
    sCM-PINN (Ours) & 2  & $\mathbf{7.52 \times 10^{-3}}$ & $1.24 \times 10^{-1}$ & $1.24 \times 10^{-1}$ \\
                    & 17 & $1.87 \times 10^{-2}$ & $2.81 \times 10^{-3}$ & $1.42 \times 10^{-4}$ \\
    \bottomrule
  \end{tabular}
  \end{newsection}
\end{table}
\end{newsection}

\begin{newsection}
\subsection{Sensitivity and Training Cost}
As shown in Table~\ref{tab:stage_transition} for the Darcy Flow benchmark, transitioning to Stage 2 at varying Stage 1 checkpoints (epochs 4, 6, and 8) yields stable metrics within the same order of magnitude. This demonstrates that our pipeline is robust to transition timing without requiring precise scheduling. Total training takes approximately 17 hours on a single NVIDIA RTX 4090 GPU. Finally, for all evaluations, we standardized our inference budget to 65 NFE for forward problems and 2 NFE for unconditional sampling.
\end{newsection}

\begin{table}[ht]
\begin{newsection}
  \centering
  \caption{Sensitivity to Stage 1 to Stage 2 transition epoch (Darcy Flow).}
  \label{tab:stage_transition}

  \renewcommand{\arraystretch}{0.8}
  \begin{tabular}{ccc}
    \toprule
    Stage 1 epoch & Relative $H^1$ (forward) & $\|\mathcal{R}\|_2 \cdot h^2$ (unconditional) \\
    \midrule
    4 & $1.08 \times 10^{-1}$ & $2.35 \times 10^{-2}$ \\
    6 & $1.09 \times 10^{-1}$ & $2.43 \times 10^{-2}$ \\
    8 & $\mathbf{1.05 \times 10^{-1}}$ & $\mathbf{2.00 \times 10^{-2}}$ \\
    \bottomrule
  \end{tabular}
\end{newsection}
\end{table}

\end{document}